\documentclass[11pt,a4paper]{article}
\usepackage[hyperref]{acl2020}
\usepackage{times}
\usepackage{latexsym}

\usepackage{microtype}

\aclfinalcopy 
\def\aclpaperid{1106} 

\usepackage{booktabs}
\usepackage{amsmath}
\usepackage{amssymb}
\usepackage{amsfonts}
\usepackage{graphicx}
\usepackage{multirow}
\usepackage{array}
\usepackage{paralist}
\usepackage[linesnumbered,ruled]{algorithm2e}

\newcommand{\citesub}[1]{\citeauthor{#1} \shortcite{#1}}
\newcommand{\RNum}[1]{\uppercase\expandafter{\romannumeral #1\relax}}

\title{Multi-Agent Task-Oriented Dialog Policy Learning with Role-Aware Reward Decomposition}

\author{Ryuichi Takanobu, Runze Liang, Minlie Huang\footnotemark[1]\\
Institute for AI, BNRist, DCST, Tsinghua University, Beijing, China\\
{\tt\small \{gxly19, liangrz15\}@mails.tsinghua.edu.cn, aihuang@tsinghua.edu.cn}}

\begin{document}
\maketitle
\renewcommand{\thefootnote}{\fnsymbol{footnote}}
\footnotetext[1]{Corresponding author}
\renewcommand{\thefootnote}{\arabic{footnote}}
\begin{abstract}
Many studies have applied reinforcement learning to train a dialog policy and show great promise these years. One common approach is to employ a user simulator to obtain a large number of simulated user experiences for reinforcement learning algorithms. However, modeling a realistic user simulator is challenging. A rule-based simulator requires heavy domain expertise for complex tasks, and a data-driven simulator requires considerable data and it is even unclear how to evaluate a simulator. To avoid explicitly building a user simulator beforehand, we propose Multi-Agent Dialog Policy Learning, which regards both the system and the user as the dialog agents. Two agents interact with each other and are jointly learned simultaneously. The method uses the actor-critic framework to facilitate pretraining and improve scalability. We also propose Hybrid Value Network for the role-aware reward decomposition to integrate role-specific domain knowledge of each agent in task-oriented dialog. Results show that our method can successfully build a system policy and a user policy simultaneously, and two agents can achieve a high task success rate through conversational interaction.
\end{abstract}

\section{Introduction}
Dialog policy, which decides the next action that the dialog agent should take, plays a vital role in a task-oriented dialog system. More recently, dialog policy learning has been widely formulated as a Reinforcement Learning (RL) problem \cite{su2016line,peng2017composite,he2018decoupling,zhao2019rethinking,zhang2019budgeted,takanobu2019guided}, which models users as the interactive environment.
Since RL requires much interaction for training, it is too time-consuming and costly to interact with real users directly. The most common way is first to develop a dialog agent with a user simulator that mimics human behaviors in an offline scenario.

Designing a reliable user simulator, however, is not trivial and often challenging as it is equivalent to building a good dialog agent. With the growing needs for the dialog system to handle more complex tasks, it will be much challenging and laborious to build a fully rule-based user simulator, which requires heavy domain expertise. Data-driven user simulators have been proposed in recent studies \cite{kreyssig2018neural,shi2019build}, but they require a considerable quantity of manually labeled data, most of which regard the simulator as a stationary environment. Furthermore, there is no standard automatic metric for evaluating these user simulators, as it is unclear to define how closely the simulator resembles real user behaviors.

In this paper, we propose Multi-Agent Dialog Policy Learning (MADPL), where the user is regarded as another dialog agent rather than a user simulator. The conversation between the user and the system is modeled as a cooperative interactive process where the system agent and the user agent are trained simultaneously. Two dialog agents interact with each other and collaborate to achieve the goal so that they require no explicit domain expertise, which helps develop a dialog system without the need of a well-built user simulator. Different from existing methods \cite{georgila2014single,papangelis2019collaborative}, our approach is based on \textit{actor-critic} framework \cite{barto1983neuronlike} in order to facilitate pretraining and bootstrap the RL training. Following the paradigm of \textit{centralized training with decentralized execution} (CTDE) \cite{bernstein2002complexity} in multi-agent RL (MARL), the actor selects its action conditioned only on its local state-action history, while the critic is trained with the actions of all agents.

\begin{figure}[!tb]
    \centering
    \includegraphics[width=\linewidth]{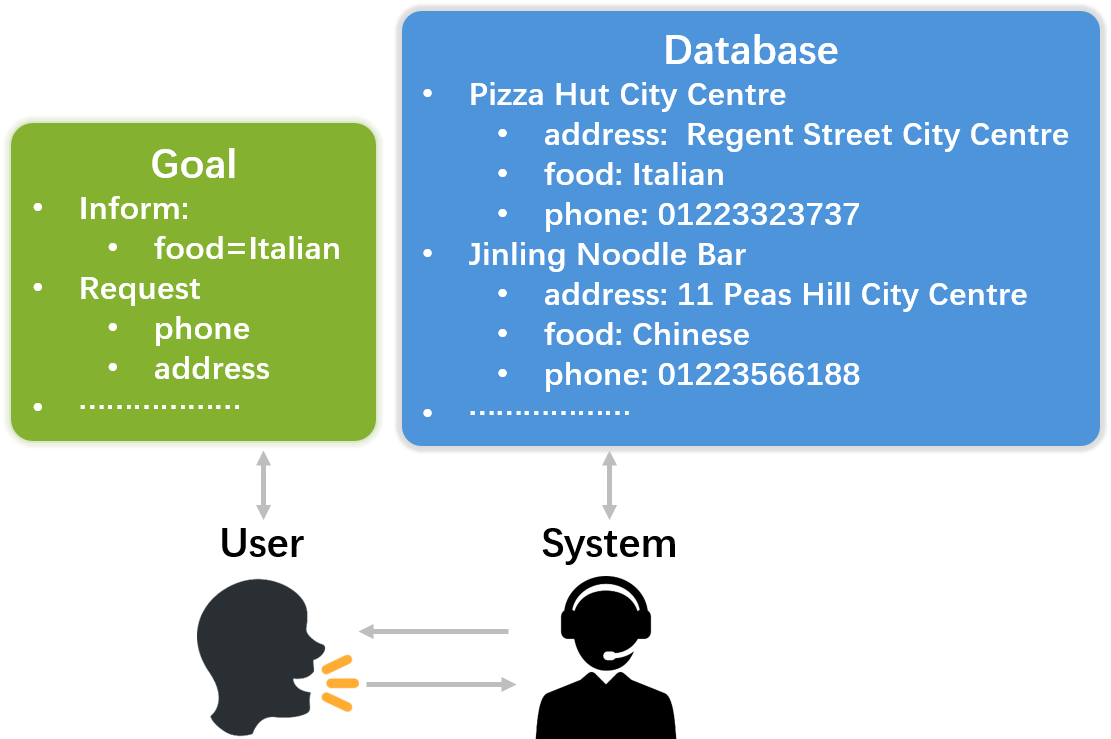} 
    \caption{The user has his/her own goal to be accomplished and the system is provided with an interface to access an external database. Both agents can only obtain information from the other side via communication.}
    \label{fig:pomdp}
\end{figure}

It should be noted that the roles of two agents are different though they interact with each other in a cooperative setting. As shown in Fig. \ref{fig:pomdp}, only the user agent knows the user goal, while only the system agent can access the backend database. The user agent should express the requirements completely in an organized way, and the system should respond with useful information accurately and immediately. So it is inappropriate to apply simple self-play RL \cite{silver2017mastering,lewis2017deal} that views two agents as the same agent in this task. To address this issue, the system and the user are viewed as two \textit{asymmetric} agents in MADPL. We introduce \textit{Hybrid Value Network} (HVN) for role-aware reward decomposition. It decomposes the reward into two parts: one is the role-specific reward that focuses on its local target, and the other is the global reward that represents the shared goal.

To evaluate the proposed approach, we conduct our experiments on a multi-domain, multi-intent task-oriented dialog corpus, MultiWOZ \cite{budzianowski2018multiwoz}. The corpus involves high dimensional state and action spaces, multiple decision making in one turn, which makes it more difficult to get a good system policy as well as a good user policy. The experiments demonstrate that MADPL can successfully build a system policy as well as a user policy with the aid of HVN, and two agents can achieve high task success rate in complex tasks by interacting with each other as well as with benchmark policies.

To summarize, our contributions are in three folds:
\begin{itemize}
    \item We apply actor-critic based multi-agent reinforcement learning to learn the task-oriented dialog policy to facilitate pretraining and avoid explicitly building a user simulator.
    \item We propose Hybrid Value Network for reward decomposition to deal with the asymmetric role issue between the system agent and the user agent in the task-oriented dialog.
    \item We conduct in-depth experiments on the multi-domain, multi-intent task-oriented dialog corpus to show the effectiveness, reasonableness and scalability of our algorithm.
\end{itemize}

\section{Related Work}
\subsection{Multi-Agent Reinforcement Learning}
The goal of RL is to discover the optimal strategy $\pi^*(a|s)$ of the Markov Decision Process, which can be extended into the $N$-agent setting, where each agent has its own set of states $\mathcal{S}_i$ and actions $\mathcal{A}_i$. In MARL, the state transition $s=(s_1, \dots, s_N) \rightarrow s'=(s'_1, \dots, s'_N)$ depends on the actions taken by all agents $(a_1, \dots, a_N)$ according to each agent's policy $\pi_i(a_i|s_i)$ where $s_i \in \mathcal{S}_i, a_i \in \mathcal{A}_i$, and similar to single RL, each agent aims to maximize its local total discounted return $R_i = \sum_{t} \gamma^t r_{i,t}$.

Since two or more agents learn simultaneously, the agents continuously change as the training proceeds, therefore the environment is no longer stationary. Many MARL algorithms \cite{lowe2017multi,foerster2018counterfactual,rashid2018qmix} have been proposed to solve challenging problems. Most of them use the CTDE framework to address the non-stationarity of co-adapting agents. It allows the policies to use extra information to ease training, but the learned policies can only use local information (i.e. their own observations) at execution time.

Several studies have demonstrated that applying MARL delivers promising results in NLP tasks these years. While some methods use identical rewards for all agents \cite{das2017learning,kottur2017natural,feng2018learning}, other studies use completely separate rewards \cite{georgila2014single,papangelis2019collaborative}. MADPL integrates two types of rewards by role-aware reward decomposition to train a better dialog policy in task-oriented dialog.

\subsection{User Modeling in Task-Oriented Dialog}
User modeling is essential for training RL-based dialog models, because a large amount of dialog samples are required for RL policy learning, making it impractical to learn with real users directly from the beginning.

There are three main approaches for user modeling.
The first approach is to build a rule-based user simulator. Among these methods, the most popular one is agenda-based simulator \cite{schatzmann2007agenda,shah2018bootstrapping}, which is built on hand-crafted rules with a stack-like agenda based on the user goal.
The second approach is to build a user simulator from the dialog data \cite{keizer2010parameter,el2016sequence,kreyssig2018neural}. Recently,
\citesub{gur2018user} uses a variational hierarchical seq2seq framework to encode user goal and system turns, and then generate the user response.
\citesub{shi2019build} uses two decoders with a copy and attention mechanism to predict a belief span first and then decode user utterance.
The third approach is to use model-based policy optimization that incorporates a differentiable model of the world dynamics and assumptions about the interactions between users and systems \cite{su2018discriminative,zhang2019budgeted}, but this approach still requires real users or a user simulator for world model learning.

Instead of employing a user simulator, a few methods jointly learn two agents directly from the corpus. \citesub{liu2017iterative} models the system and the user by iteratively training two policies.
\citesub{papangelis2019collaborative} make the first attempt to apply MARL into the task-oriented dialog policy, whose algorithm is based on Q-learning for mixed policies. However, it is not well scalable to complex tasks such as multi-domain dialog. Therefore, MADPL uses the actor-critic framework instead to deal with the large discrete action space in dialog.

\section{Multi-Agent Dialog Policy Learning}

We first formally describe the task, and then present the overview of our proposed model. Specifically, given a user goal \texttt{G}=(\texttt{C},\texttt{R}) composed of the user constraints \texttt{C} (e.g. a Japanese restaurant in the center of the city) and requests \texttt{R} (e.g. inquiry for address, phone number of a hotel), and given an external database \texttt{DB} containing all candidate entities and corresponding information, the user agent and system agent interact with each other in a dialog session to fulfill the user goal. There can be multiple domains in \texttt{G}, and two agents have to accomplish all the subtasks in each domain. Both agents can partially observe the environment, i.e. only the user agent knows \texttt{G}, while only the system agent can access \texttt{DB}, and the only way to know each other's information is through conversational interaction. Different from ordinary multi-agent task setting, two agents in dialog are executed \textit{asynchronously}. In a single dialog turn, the user agent posts an inquiry first, then the system agent returns a response, and the two communicate alternately. Therefore, each dialog session $\tau$ can be seen as a trajectory of state-action pairs $\{(s_0^U, a_0^U, s_0^S, a_0^S); (s_1^U, a_1^U, s_1^S, a_1^S);\dots\}$, where the user agent and the system agent make decisions according to each dialog policy $\mu(a^U|s^U), \pi(a^S|s^S)$ respectively.

\begin{figure}[!tb]
    \centering
    \includegraphics[width=\linewidth]{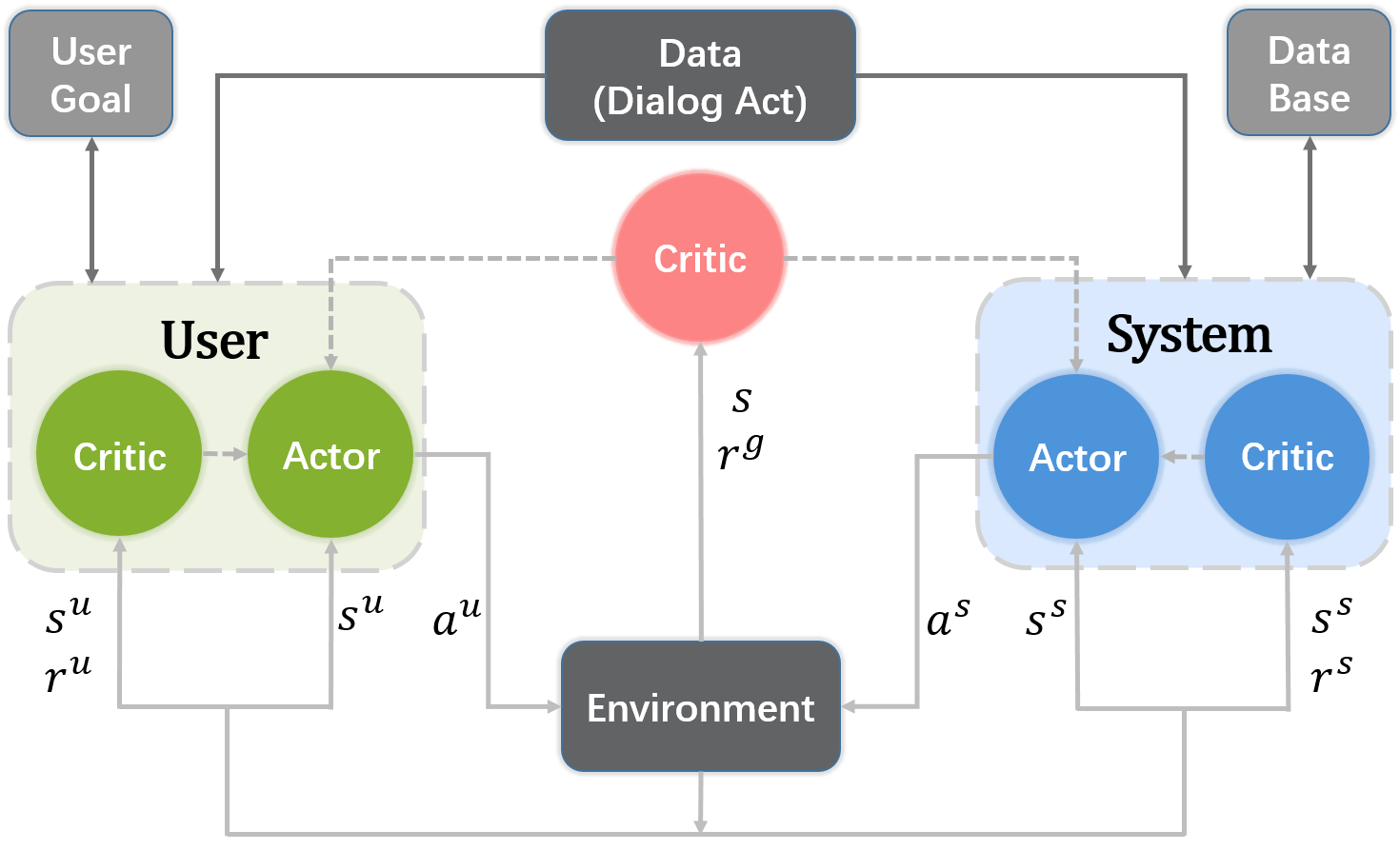}
    \caption{Architecture of MADPL. HVN consists of three critics. Each critic estimates its return based on role-aware reward decomposition, and each actor uses the estimated value to optimize itself.}
    \label{fig:madpl}
\end{figure}

Here we present a novel algorithm, Multi-Agent Dialog Policy Learning (MADPL), as shown in Fig. \ref{fig:madpl}, which can be naturally formulated as a MARL problem. Two agents interact through dialog acts following \cite{georgila2014single}. We choose the \textit{actor-critic} framework in order to learn an explicitly stochastic dialog policy (actor) for high scalability along with an estimated value function (critic) to bootstrap RL training. Besides, this can facilitate imitation learning to pretrain the dialog policy using human-human dialogs. Since two agents cooperate to reach success, yet their roles are asymmetric in the dialog, we propose Hybrid Value Network (HVN) to decompose the task reward into different parts for better policy learning. Note that our approach is fully data-driven without building a user simulator beforehand, and does not need any other human supervision during training.

In the subsequent subsections, we will first explain the state and action used in two dialog policies. Then we describe how we decompose the reward and the proposed HVN. At last, we present model optimization.

\subsection{Dialog Policy}

\paragraph{System Policy}
The system policy $\pi$ decides the system action $a^S$ according to the system dialog state $s^S$ to give the appropriate response to user agent. Each system action $a^S$ is a subset of dialog act set $\mathcal{A}$ as there may be multiple intents in one dialog turn. A \textit{dialog act} is an abstract representation of an intention \cite{stolcke2000dialogue}, which can be represented in a quadruple composed of \textit{domain}, \textit{intent}, \textit{slot type} and \textit{slot value} (e.g. [\texttt{restaurant}, \texttt{inform}, \texttt{food}, \texttt{Italian}]). In practice, dialog acts are \textit{delexicalized} in the dialog policy. We replace the slot value with a count placeholder and refill it with the true value according to the entity selected from the external database \texttt{DB}, which allows the system to operate on unseen values. The system dialog state $s^S_t$ at dialog turn $t$ is the concatenation of (\RNum{1}) user action at current turn $a_t^U$; (\RNum{2}) system action at the last turn $a_{t-1}^U$; (\RNum{3}) the belief state $b_t$ \cite{williams2016dialog} that keeps track of constraint slots and request slots supplied by the user agent; and (\RNum{4}) embedding vectors of the number of query results $q_t$ from \texttt{DB}.

\paragraph{User Policy}
The user policy $\mu$ decides the user action $a^U$ according to the user dialog state $s^U$ to express its constraint and request to the system agent. Similar to the system policy, the user policy uses delexicalized dialog acts as actions, and the value is refilled according to the user goal \texttt{G}. User dialog state $s^U_t$ is the concatenation of (\RNum{1}) last system action $a_{t-1}^S$; (\RNum{2}) last user action $a_{t-1}^U$; (\RNum{3}) the goal state $g_t$ that represents the remained constraint and request that need to send; (\RNum{4}) inconsistency vector $c_t$ \cite{kreyssig2018neural} that indicates the inconsistency between the system’s response and user constraint \texttt{C}. In addition to predicting dialog acts, the user policy outputs terminal signal $T$ at the same time, i.e. $\mu = \mu(a^U,T|s^U)$.

\subsection{Reward Decomposition}

On the one hand, the roles between the user agent and the system agent are different. The user agent actively initiates a task and may change it during conversation, but the system agent passively responds to the user agent and returns the proper information, so the reward should be considered separately for each agent. On the other hand, two agents communicate and collaborate to accomplish the same task cooperatively, so the reward also involves a global target for both agents. Therefore, we decompose the mixed reward into three parts according to the characteristic of each component. The reward of each part is explained as follows:

\paragraph{System Reward} $r_t^S$ consists of (\RNum{1}) empty dialog act penalty $a_t^S = \varnothing$; (\RNum{2}) late answer penalty if there is a request slot triggered but the system agent does not reply the information immediately; and (\RNum{3}) task success reward based on the user agent's description.

\paragraph{User Reward} $r_t^U$ consists of (\RNum{1}) empty dialog act penalty $a_t^U = \varnothing$; (\RNum{2}) early inform penalty if the user agent requests for information when there is still a constraint slot remained to inform; and (\RNum{3}) user goal reward whether the user agents have expressed all the constraints \texttt{C} and requests \texttt{R}.

\paragraph{Global Reward} $r_t^G$ consists of (\RNum{1}) efficiency penalty that a small negative value will be given at each dialog turn; (\RNum{2}) sub-goal completion reward once the subtask of \texttt{G} in a particular domain is accomplished; and (\RNum{3}) task success reward based on user goal \texttt{G}.

Obviously, each agent should obtain its local reward, and both agents should receive the global reward during the training process.
Note that the task success and the user goal reward are only computed at the end of the dialog, and the task success computed in the system reward differs from the one in the global reward.

\subsection{Hybrid Value Network}
The value function aims to estimate the expected return given the current state $V(s_t) = \mathbb{E}[R_t] = \mathbb{E}[\sum_{t'\geq t}\gamma^{t'-t}r_{t'}]$ so that the policy can directly use the estimated cumulative reward for optimization, without sampling the trajectories to obtain rewards which may cause high variance.
Another advantage by applying actor-critic approaches in MARL is that it can integrate with the CTDE framework: the actor of each agent benefits from a critic that is augmented with additional information about the policies of other agents during training. However, a simple centralized critic conditioned on the global state and joint actions cannot well exploit the domain knowledge mentioned above since each part of the overall rewards only depends on a subset of features, e.g. the system reward only depends on the system agent's behaviors.

Inspired by \textit{Hybrid Reward Architecture} \cite{van2017hybrid} that learns a separate Q function, we propose Hybrid Value Network to improve an estimate of the optimal role-aware value function. It first encodes the dialog state of each agent to learn a state representation
\begin{align*}
    h_s^S &= \tanh(f^S_s(s^S)), \\
    h_s^U &= \tanh(f^U_s(s^U)),
\end{align*}
where $f(\cdot)$ can be any neural network unit. The value network $V$ is separated into three branches $V^S$, $V^U$ and $V^G$ for the value of system rewards, user rewards and global rewards, respectively.
\begin{align*}
    V^S(s^S) &= f_S(h_s^S), \\
    V^U(s^U) &= f_U(h_s^U), \\
    V^G(s) &= f_G([h_s^S;h_s^U]).
\end{align*}

\subsection{Optimization}
The action space for the policies can be very large since we deal with multi-domain, complex dialog tasks, which makes it almost impossible for the RL policies to explore and learn from scratch. So the training process can be split into two stages \cite{fatemi2016policy,takanobu2019guided}: pretraining the dialog policy with the conversational corpus first and then using RL to improve the pretrained policies.
We use $\beta$-weighted logistic regression for policy pretraining here to alleviate data bias because each agent only generates several dialog acts in one dialog turn
\begin{align}\label{eq:pretrain}
    L(X, Y; \beta) =& -[\beta \cdot Y^T \log\sigma(X)\\
    &+ (I-Y)^T\log(I -\sigma(X))], \notag
\end{align}
where $X$ is the state and $Y$ is the action from the corpus in this task.

As for critic optimization, it aims to minimize the squared error between the temporal difference (TD) target $r_t + \gamma V(s_{t+1})$ and the estimated value $V(s_t) = \mathbb{E}[r_t + \gamma V(s_{t+1})]$. Actor-critic algorithms have high variance  since the critic is updated too frequently, which has contributed to severe changes in the estimated value, particularly in multi-agent tasks. So we introduce a target network \cite{mnih2015human} to make the training process more stable. In the context of HVN, it aims to minimize the following loss functions:
\begin{align}\label{eq:value_network}
    L_V^S(\theta)
    &= (r^S + \gamma V^S_{\theta^-}(s'^S) - V^S_\theta(s^S))^2, \notag \\
    L_V^U(\theta)
    &= (r^U + \gamma V^U_{\theta^-}(s'^U) - V^U_\theta(s^U))^2, \notag \\
    L_V^G(\theta)
    &= (r^G + \gamma V^G_{\theta^-}(s') - V^G_\theta(s))^2, \notag \\
    L_V &= L_V^S + L_V^U + L_V^G,
\end{align}
where HVN $V_\theta$ is parameterized by $\theta$, and $\theta^-$ is the weight of target network, and the overall loss $L_V$ is the sum of value estimation loss on each component reward.

\begin{algorithm}[!tb]
\DontPrintSemicolon
\caption{Multi-Agent Dialog Policy Learning}
\label{algorithm}
\SetKwInOut{Input}{Require}
\Input{Dialog corpus $\mathcal{D}$ with annotations of dialog acts \{$a$\}}
Initialize weights $\phi, \omega$ for system policy $\pi$ and user policy $\mu$ respectively \\
Pretrain policies $\pi, \mu$ on human conversational data $\mathcal{D}$ using Eq. \ref{eq:pretrain} \\
Initialize weights $\theta$ for hybrid value network $V=(V^S, V^U, V^G)$ and target network $\theta^- \leftarrow \theta$ \\
\ForEach{training iteration}{
    Initialize user goal and dialog state $s^U, s^S$ \\
    \Repeat{the session ends according to $T$}{
        Sample actions $a^U, a^S$ and terminal signal $T$ using current policy $\pi, \mu$ \\
        Execute actions and observe reward $r^U, r^S, r^G$ and new states $s'^U, s'^S$ \\
        Update hybrid value network (critic) using Eq. \ref{eq:value_network} \\
        Compute the advantage $A^U, A^S, A^G$ using current value network \\
        Update two dialog policies (actor) using Eq. \ref{eq:policy_network} \\
        $s^U \leftarrow s'^U, s^S \leftarrow s'^S$ \\
        Assign target network parameters $\theta^- \leftarrow \theta$ every $C$ steps
    }
}
\end{algorithm}

Each dialog policy aims to maximize all the related returns, e.g. the system policy $\pi$ aims to maximize the cumulative system rewards and global rewards $\mathbb{E}[\sum_t \gamma^t(r_t^S + r_t^G)]$. The advantage $A(s)=r+\gamma V(s')-V(s)$ estimated by the critic can evaluate the new state $s'$ and current state $s$ to determine whether the dialog has become better or worse than expected. With the aid of HVN, the sum of the related component advantages can be used to update different agents. By using the log-likelihood ratio trick, the gradients for the system policy and the user policy yield:
\begin{align}\label{eq:policy_network}
    \nabla_{\phi} J_{\pi}(\phi) = \nabla_{\phi} \log \pi_\phi(a^S|s^S) [A^S(s^S) + A^G(s)], \\
    \nabla_{\omega} J_{\mu}(\omega) = \nabla_{\omega} \log \mu_\omega(a^U|s^U) [A^U(s^U) + A^G(s)], \notag
\end{align}
where the system policy $\pi_\phi$ is parameterized by $\phi$ and the user policy $\mu_\omega$ by $\omega$.

In summary, a brief script for MADPL is shown in Algorithm \ref{algorithm}.

\section{Experimental Setting}

\subsection{Dataset}
MultiWOZ \cite{budzianowski2018multiwoz} is a multi-domain, multi-intent task-oriented dialog corpus that contains 7 domains, 13 intents, 25 slot types, 10,483 dialog sessions, and 71,544 dialog turns. During the data collection process, a user is asked to follow a pre-specified user goal, and is allowed to change the goal during the session if necessary, so the collected dialogs are much closer to real-world conversations. The corpus also provides the domain knowledge that defines all the entities and attributes as the external database.
\subsection{Metrics}
Evaluation of a task-oriented dialog system mainly consists of the cost and task success. We count the number of \textit{dialog turns} to reflect the dialog cost. A user utterance and a subsequent system utterance are regarded as one dialog turn. We utilize two other metrics: \textit{inform F1} and \textit{match rate} to estimate the task success. Both metrics are calculated at the dialog act level.
Inform F1 evaluates whether all the \textit{requested information} has been informed, and match rate checks whether the booked entities match all the \textit{indicated constraints} given by the user. The overall task success is reached if and only if both inform recall and match rate are 1.

\subsection{Baselines}

We compare MADPL with a series of baselines that involve both system policy learning and user policy learning. Note that we do not consider any other approaches that use a user simulator for policy training because our motivation is to avoid explicitly modeling a simulator.

\paragraph{SL}  \textit{Supervised Imitation Learning} directly uses the dialog act annotations and trains the agents simply by behavior cloning using Eq. \ref{eq:pretrain}, which is the same as the pretraining phase in MADPL.

The following three baselines are all RL algorithms that start from the pretrained policy:

\paragraph{RL} \textit{Independent Reinforcement Learning} learns only one dialog policy by fixing another agent following the single RL setting, and the reward for the agent is the sum of role-specific reward and global reward. For example, the user policy uses the reward $r = r^U + r^G$ at each dialog turn.

\paragraph{CRL} \textit{Centralized Reinforcement Learning} is a MARL approach that uses a single centralized critic on the sum of reward $r = r^U + r^S + r^G$ to train two agents simultaneously, which also serves for the ablation test of MADPL.

\paragraph{IterDPL} \textit{Iterative Dialog Policy Learning} \cite{liu2017iterative} updates two agents iteratively using single RL training to reduce the risk of non-stationarity when jointly training the two agents.

\begin{table}[!tb]
    \centering
    \small
    \begin{tabular}{cccc}
    \toprule
        Class & Attraction & Hospital & Hotel \\
        Count & 320 & 22 & 389 \\
    \cmidrule(lr){0-3}
        Police & Restaurant & Taxi & Train \\
        22 & 457 & 164 & 421 \\
    \midrule
        Num. & Single & Two & Three \\
        Count & 328 & 549 & 123 \\
    \bottomrule
    \end{tabular}
    \caption{Domain distribution of user goals used in the automatic evaluation. A user goal with multiple domains is counted repeatedly for each domain.}
    \label{tab:goal}
\end{table}

\section{Automatic Evaluation}

\subsection{Interaction between Two Agents}

\begin{figure}[!tb]
    \centering
    \includegraphics[width=\linewidth]{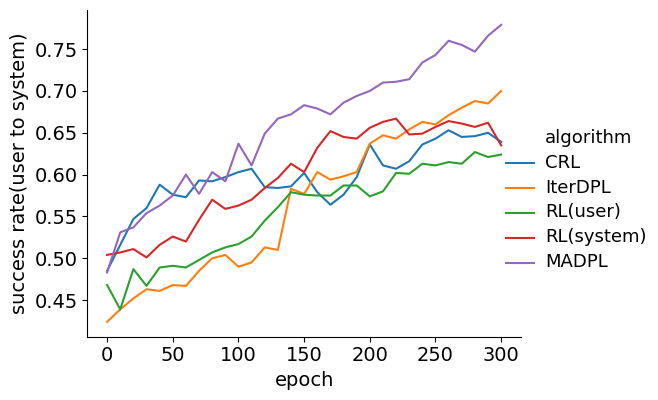}
    \caption{Learning curves of the interaction between the user agent and the system agent.}
    \label{fig:2agent}
\end{figure}

A set of 1,000 user goals are used for automatic evaluation as shown in Table \ref{tab:goal}. When the dialog is launched, two agents interact with each other around a given user goal.
The performance of interaction between the two trained policies are shown in Table \ref{tab:2agent}. MADPL reaches the highest match rate and task success among all the methods. It manages to improve the success rate of the pretrained policies from 49.7\% to 70.1\%. Single RL policies (row 2 to 4) have limited improvement, and even decline in match rate since they assume a stationary environment. The comparison between CRL and IterDPL indicates the effectiveness of CTDE in the multi-agent task. The superiority of MADPL against CRL shows that two agents benefit from the role-aware reward decomposition in HVN. The learning curves in Fig. \ref{fig:2agent} illustrates that the success rate grows rapidly in MADPL, and it always improves the success rate as the training proceeds.

\begin{figure}[!tb]
    \centering
    \includegraphics[width=0.81\linewidth]{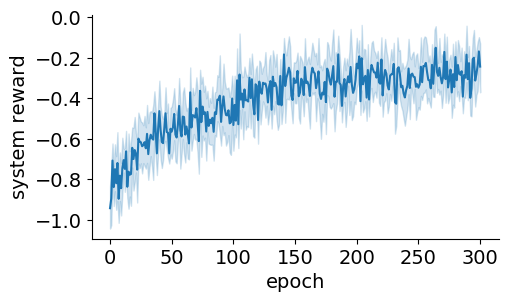}
    \includegraphics[width=0.81\linewidth]{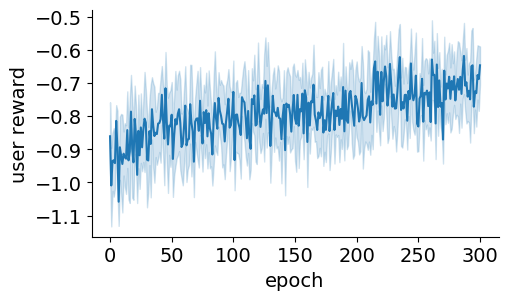}
    \includegraphics[width=0.81\linewidth]{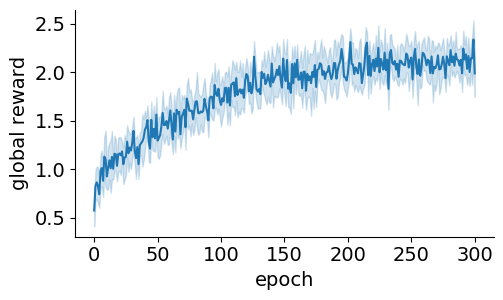}
    \caption{Learning curves of MADPL on system reward (top), user reward (middle) and global reward (bottom).}
    \label{fig:reward}
\end{figure}

The average reward of each component reward is shown in \ref{fig:reward}. We run 10 different instances of MADPL with different random seeds. The solid curves correspond to the mean and the shaded region to the standard deviation of rewards over the 10 trials. We can observe that all the rewards increase steadily during the training process, which implies that HVN has estimated a proper return for policy training.

\begin{table}[!tb]
    \centering
    \begin{tabular}{c@{~~}c@{~~}|@{~~}c@{~~}c@{~~}c@{~~}c}
    \toprule
        System & User & Turns & Inform & Match & Success \\
    \midrule
        SL & SL & 6.34 & 73.08 & 82.58 & 49.7 \\
        SL & RL & 8.75 & \textbf{76.86} & 76.28 & 60.2 \\
        RL & SL & 6.20 & 72.84 & 79.15 & 51.1 \\
        RL & RL & 7.92 & 75.96 & 70.37 & 58.7\\
        \multicolumn{2}{c|@{~~}}{CRL} & 8.13 & 68.29 & 89.71 & 66.6 \\
        \multicolumn{2}{c|@{~~}}{IterDPL} & 8.79 & 74.01 & 81.04 & 64.6 \\
    \midrule
        \multicolumn{2}{c|@{~~}}{MADPL} & 8.96 & 76.26 & \textbf{90.98} & \textbf{70.1} \\
    \bottomrule
    \end{tabular}
    \caption{Performance of the interaction between the user agent and the system agent.}
    \label{tab:2agent}
\end{table}

\subsection{Interaction with Benchmark Policies}

It is essential to evaluate a multi-agent dialog system whether all the agents understand the semantic interaction rather than invent an uninterpretable language \cite{kottur2017natural,lee2019countering}. To this end, we use two benchmark policies in the standardized task-oriented dialog system platform Convlab \cite{lee2019convlab} to examine all the methods. Each benchmark is a strong rule-based system policy or user policy at the dialog act level, which is used as the simulated evaluation in the DSTC-8 Track 1 competition and show a high correlation with real user interaction \cite{li2020results}.
The trained system/user policy in each method is directly deployed to interact with the benchmark user/system policy during the test without any other finetuning, which can be regarded as a weakly zero-shot experiment. The same goal set in Table \ref{tab:goal} is used here.

\begin{figure}[!tb]
    \centering
    \includegraphics[width=\linewidth]{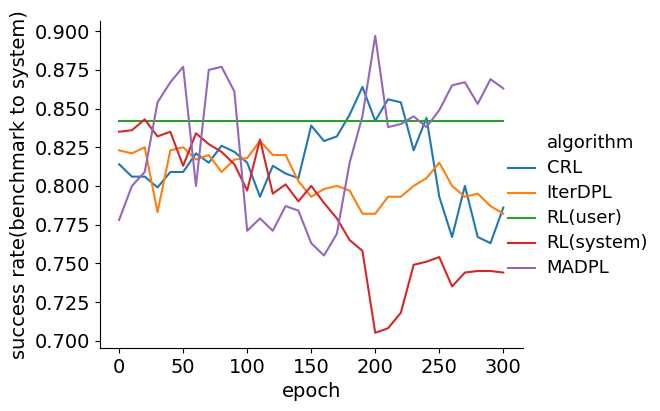}
    \caption{Learning curves of the interaction between the benchmark user policy and each system agent.}
    \label{fig:rule_system}
\end{figure}

Table \ref{tab:rule_system} and Fig. \ref{fig:rule_system} show the results of the interaction between the benchmark user policy and the system agent of each model. The SOTA performance from GDPL \cite{takanobu2019guided} that directly trains with benchmark user policy is also presented as the soft performance upper bound. Among all the methods, MADPL has achieved the highest task success and the second-highest match rate. All the methods experience a decline in inform F1 after the RL training. Fig. \ref{fig:rule_system} also shows that the success rate is unstable during training. This is because the action space of the system policy is much larger, thus more challenging to learn. In spite of that, the success rate of MADPL shows a rising trend.

\begin{table}[!tb]
    \centering
    \begin{tabular}{c|cccc}
    \toprule
        System & Turns & Inform & Match & Success \\
    \midrule
        SL & 7.76 & \textbf{83.33} & 85.84 & 84.2 \\
        RL & 7.53 & 82.06 & 85.77 & 84.3\\
        CRL & 8.38 & 72.43 & \textbf{89.48} & 86.4 \\
        IterDPL & 7.74 & 79.68 & 82.49 & 82.5 \\
    \midrule
        MADPL & 7.63 & 79.93 & 89.24 & \textbf{87.7} \\
        GDPL & \textit{7.62} & \textit{92.10} & \textit{91.50} & \textit{92.1} \\
    \bottomrule
    \end{tabular}
    \caption{Performance of the interaction between the benchmark user policy and each system agent.}
    \label{tab:rule_system}
\end{table}

\begin{table}[!tb]
    \centering
    \begin{tabular}{c|cccc}
    \toprule
        User & Turns & Inform & Match & Success \\
    \midrule
        SL & 8.64 & 78.64 & 87.84 & 51.7 \\
        RL & 11.18 & 85.69 & 92.13 & 77.2 \\
        CRL & 11.31 & 86.58 & \textbf{92.89} & 74.7 \\
        IterDPL & 12.53 & 84.68 & 92.57 & 75.5\\
    \midrule
        MADPL & 13.25 & \textbf{87.04} & 90.81 & \textbf{83.7} \\
    \bottomrule
    \end{tabular}
    \caption{Performance of the interaction between each user agent and the benchmark system policy.}
    \label{tab:rule_user}
\end{table}

\begin{figure}[!tb]
    \centering
    \includegraphics[width=\linewidth]{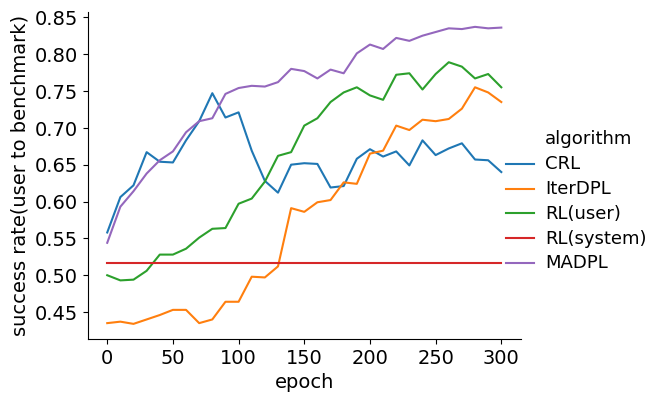}
    \caption{Learning curves of the interaction between each user agent and the benchmark system policy.}
    \label{fig:rule_user}
\end{figure}

\begin{figure*}[!tb]
    \centering
    \includegraphics[width=0.40\linewidth]{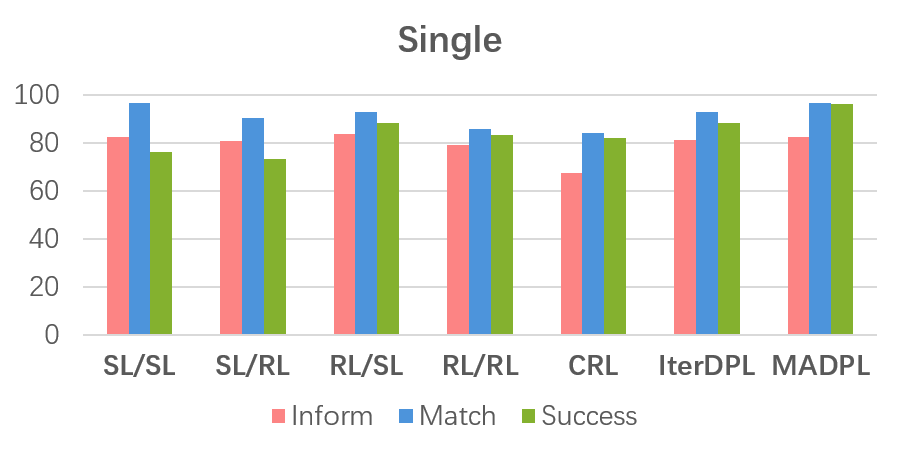}
    \includegraphics[width=0.40\linewidth]{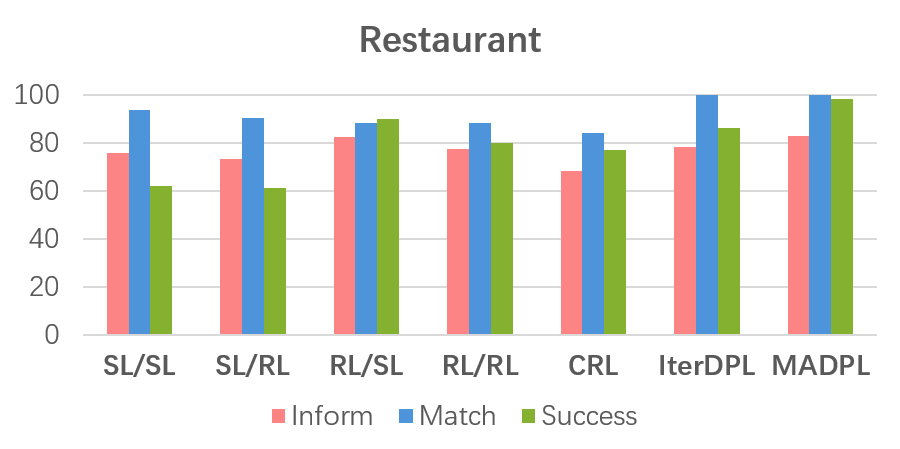}
    \includegraphics[width=0.40\linewidth]{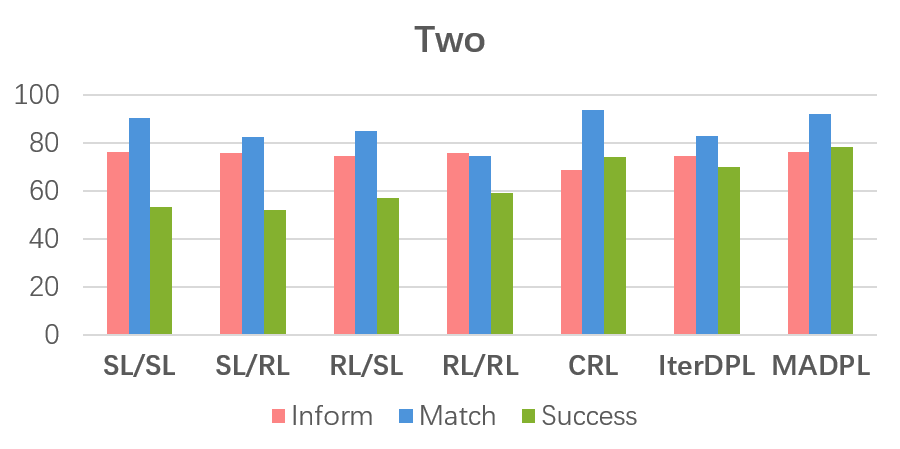}
    \includegraphics[width=0.40\linewidth]{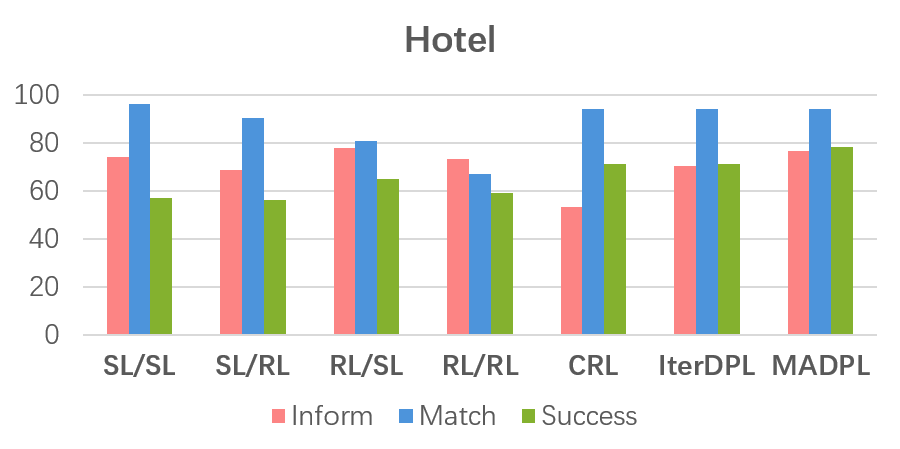}
    \includegraphics[width=0.40\linewidth]{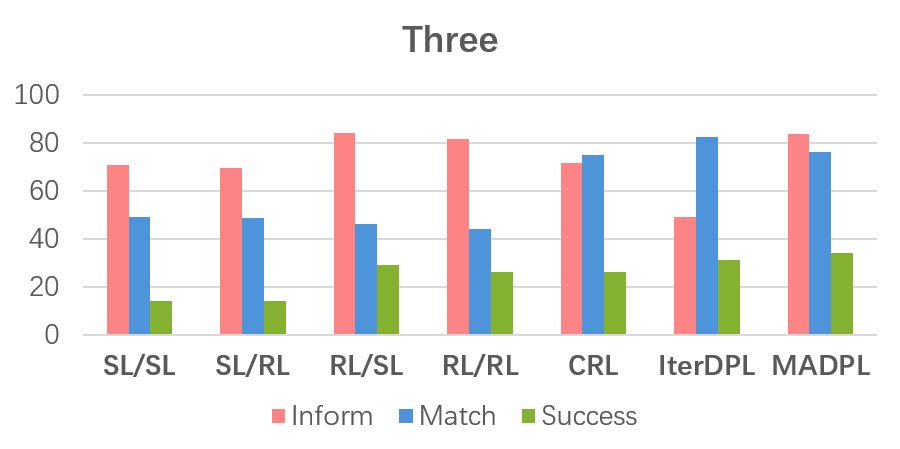}
    \includegraphics[width=0.40\linewidth]{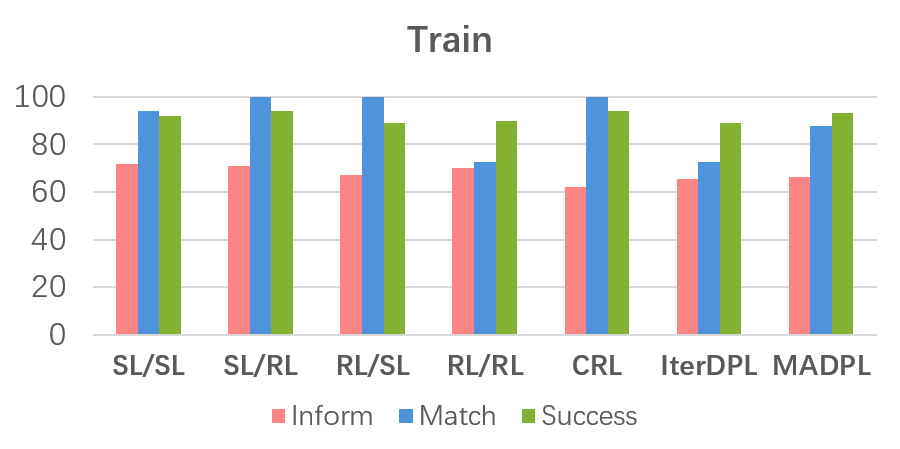}
    \caption{Performance of dialog agents according to the different number (left) or class (right) of domains in the dialog.}
    \label{fig:domain}
\end{figure*}

Table \ref{tab:rule_user} and Fig. \ref{fig:rule_user} show the results of the interaction between the user agent of each method and the benchmark system policy. Among all the methods, MADPL has achieved the highest inform F1 and task success. Though CRL improves the performance at the beginning, the success rate fails to increase further afterwards, while MADPL continues to improve all the time. This also indirectly indicates the advantage of using role-aware reward decomposition in HVN.

In summary, each policy trained from MADPL can interact well with the benchmark policy, which implies that MADPL learns a reasonable dialog strategy.

\subsection{Goal across Multiple Domains}

We also investigate the domains in the user goals to observe the scalability of each method in the complex tasks. 200 goals are randomly sampled under each setting. Fig. \ref{fig:domain} presents the results of the interaction between two agents in different numbers or classes of domains. The success rate decreases substantially as the number of domains increases in the goal. When there are 3 domains in the goal, RL/RL gets a high inform F1 but a low match rate, IterDPL gets a high match rate but a low inform F1, while MADPL can still keep a high inform F1 and match rate, and obtains the highest task success. In terms of the class of domains, there are 7/10/6 informable slots that needs to be tracked in the \textit{Restaurant}/\textit{Hotel}/\textit{Train} domain respectively. Among these, MADPL outperforms other baselines in the \textit{Restaurant} and \textit{Hotel} domains, and performs comparably in the \textit{Train} domain. In brief, all the results indicate that MADPL has good scalability in multi-domain dialog.

\section{Human Evaluation}

\begin{table}[!tb]
    \centering
    \begin{tabular}{l@{~~}c@{~~~}c@{~~~}c@{~~~}c@{~~~}c@{~~~}c@{~~~}c@{~~~}c@{~~~}c}
    \toprule
        \multirow{2}{*}[-0.04in]{VS.} & \multicolumn{3}{@{~}l@{}}{System Q} & \multicolumn{3}{@{~~~}l@{}}{User Q} & \multicolumn{3}{@{~~}l@{}}{Success} \\
        \cmidrule(l{-0.1em}r{0.8em}){2-4} \cmidrule(l{-0.1em}r{0.8em}){5-7} \cmidrule(l{-0.1em}r{0.7em}){8-10}
        & W & D & L & W & D & L & W & D & L\\
    \midrule
        SL/SL & 55 & 22 & 23 & 61 & 25 & 14 & 68 & 26 & 6 \\
        RL/RL & 49 & 23 & 28 & 52 & 28 & 20 & 70 & 19 & 11 \\
        IterDPL & 50 & 27 & 23 & 56 & 30 & 14 & 64 & 24 & 12 \\
    \bottomrule
    \end{tabular}
    \caption{Human preference on dialog session pairs that MADPL wins (W), draws with (D) or loses to (L) baselines with regard to quality (Q) and success by majority voting.}
    \label{tab:human}
\end{table}

For human evaluation, we hire Amazon Mechanical Turkers to conduct pairwise comparison between MADPL and baselines. Since all the policies work at the dialog act level, we generate the texts from dialog acts using hand-crafted templates to make the dialog readable. Each Turker is asked to read a user goal first, then we show 2 dialog sessions around this user goal, one from MADPL and the other from another baseline. We randomly sample 100 goals for each baseline. For each goal, 5 Turkers are asked to judge which dialog is better (win, draw or lose) according to different subjective assessments independently: (\RNum{1}) system quality, (\RNum{2}) user quality, and (\RNum{3}) task success. The system quality metric evaluates whether the system policy provides the user with the required information efficiently, and the user quality metric evaluates whether the user policy expresses the constraints completely in an organized way. Note that we do not evaluate the quality of language generation here. 

Table \ref{tab:human} shows the results of human preference by majority voting. We can observe that the high win rate of MADPL on the task success is consistent with the results of automatic evaluation, and MADPL outperforms three baselines significantly in all aspects (sign test, p-value $<$ 0.01) except for the system quality against RL/RL policies.

The proportion of the pairwise annotations in which at least 3 of 5 annotators assign the same label to a task is 78.7\%/77.3\%/83.3\% for system quality/user quality/task success, respectively. This indicates that annotators have moderate agreements. The human judgements align well with the results of automatic evaluation, which also indicates the reliability of the metrics used in task-oriented dialog.

\section{Conclusion}

We present a multi-agent dialog policy algorithm, MADPL, that trains the user policy and the system policy simultaneously. It uses the actor-critic framework to facilitate pretraining and bootstrap RL training in multi-domain task-oriented dialog. We also introduce role-aware reward decomposition to integrate the task knowledge into the algorithm.
MADPL enables the developers to set up a dialog system rapidly from scratch. It only requires the annotation of dialog acts in the corpus for pretraining and does not need to build a user simulator explicitly beforehand.
Extensive experiments\footnote{We provide implementation details and case studies in appendix.} demonstrate the effectiveness, reasonableness and scalability of MADPL.

As future work, we will apply MADPL in the more complex dialogs and verify the role-aware reward decomposition in other dialog scenarios.

\section*{Acknowledgement}
This work was jointly supported by the NSFC projects (Key project with No. 61936010 and regular project with No. 61876096), and the National Key R\&D Program of China (Grant No. 2018YFC0830200). We would like to thank THUNUS NExT Joint-Lab for the support.
The code is available at \url{https://github.com/truthless11/MADPL}.

\bibliography{acl2020}

\begin{thebibliography}{36}
\expandafter\ifx\csname natexlab\endcsname\relax\def\natexlab#1{#1}\fi

\bibitem[{Barto et~al.(1983)Barto, Sutton, and Anderson}]{barto1983neuronlike}
Andrew~G Barto, Richard~S Sutton, and Charles~W Anderson. 1983.
\newblock Neuronlike adaptive elements that can solve difficult learning
  control problems.
\newblock \emph{IEEE transactions on systems, man, and cybernetics},
  13(5):834--846.

\bibitem[{Bernstein et~al.(2002)Bernstein, Givan, Immerman, and
  Zilberstein}]{bernstein2002complexity}
Daniel~S Bernstein, Robert Givan, Neil Immerman, and Shlomo Zilberstein. 2002.
\newblock The complexity of decentralized control of markov decision processes.
\newblock \emph{Mathematics of operations research}, 27(4):819--840.

\bibitem[{Budzianowski et~al.(2018)Budzianowski, Wen, Tseng, Casanueva, Ultes,
  Ramadan, and Ga{\v{s}}i{\'c}}]{budzianowski2018multiwoz}
Pawe{\l} Budzianowski, Tsung-Hsien Wen, Bo-Hsiang Tseng, I{\~n}igo Casanueva,
  Stefan Ultes, Osman Ramadan, and Milica Ga{\v{s}}i{\'c}. 2018.
\newblock Multiwoz: A large-scale multi-domain wizard-of-oz dataset for
  task-oriented dialogue modelling.
\newblock In \emph{2018 Conference on Empirical Methods in Natural Language
  Processing}, pages 5016--5026.

\bibitem[{Das et~al.(2017)Das, Kottur, Moura, Lee, and Batra}]{das2017learning}
Abhishek Das, Satwik Kottur, Jos{\'e}~MF Moura, Stefan Lee, and Dhruv Batra.
  2017.
\newblock Learning cooperative visual dialog agents with deep reinforcement
  learning.
\newblock In \emph{2017 IEEE International Conference on Computer Vision},
  pages 2951--2960.

\bibitem[{El~Asri et~al.(2016)El~Asri, He, and Suleman}]{el2016sequence}
Layla El~Asri, Jing He, and Kaheer Suleman. 2016.
\newblock A sequence-to-sequence model for user simulation in spoken dialogue
  systems.
\newblock \emph{17th Annual Conference of the International Speech
  Communication Association}, pages 1151--1155.

\bibitem[{Fatemi et~al.(2016)Fatemi, El~Asri, Schulz, He, and
  Suleman}]{fatemi2016policy}
Mehdi Fatemi, Layla El~Asri, Hannes Schulz, Jing He, and Kaheer Suleman. 2016.
\newblock Policy networks with two-stage training for dialogue systems.
\newblock In \emph{17th Annual Meeting of the Special Interest Group on
  Discourse and Dialogue}, pages 101--110.

\bibitem[{Feng et~al.(2018)Feng, Li, Huang, Liu, Ou, Wang, and
  Zhu}]{feng2018learning}
Jun Feng, Heng Li, Minlie Huang, Shichen Liu, Wenwu Ou, Zhirong Wang, and
  Xiaoyan Zhu. 2018.
\newblock Learning to collaborate: Multi-scenario ranking via multi-agent
  reinforcement learning.
\newblock In \emph{27th International Conference on World Wide Web}, pages
  1939--1948.

\bibitem[{Foerster et~al.(2018)Foerster, Farquhar, Afouras, Nardelli, and
  Whiteson}]{foerster2018counterfactual}
Jakob~N Foerster, Gregory Farquhar, Triantafyllos Afouras, Nantas Nardelli, and
  Shimon Whiteson. 2018.
\newblock Counterfactual multi-agent policy gradients.
\newblock In \emph{32nd AAAI Conference on Artificial Intelligence}, pages
  2974--2982.

\bibitem[{Georgila et~al.(2014)Georgila, Nelson, and
  Traum}]{georgila2014single}
Kallirroi Georgila, Claire Nelson, and David Traum. 2014.
\newblock Single-agent vs. multi-agent techniques for concurrent reinforcement
  learning of negotiation dialogue policies.
\newblock In \emph{52nd Annual Meeting of the Association for Computational
  Linguistics}, pages 500--510.

\bibitem[{G{\"u}r et~al.(2018)G{\"u}r, Hakkani-T{\"u}r, T{\"u}r, and
  Shah}]{gur2018user}
Izzeddin G{\"u}r, Dilek Hakkani-T{\"u}r, Gokhan T{\"u}r, and Pararth Shah.
  2018.
\newblock User modeling for task oriented dialogues.
\newblock In \emph{2018 IEEE Spoken Language Technology Workshop}, pages
  900--906.

\bibitem[{He et~al.(2018)He, Chen, Balakrishnan, and Liang}]{he2018decoupling}
He~He, Derek Chen, Anusha Balakrishnan, and Percy Liang. 2018.
\newblock Decoupling strategy and generation in negotiation dialogues.
\newblock In \emph{2018 Conference on Empirical Methods in Natural Language
  Processing}, pages 2333--2343.

\bibitem[{Keizer et~al.(2010)Keizer, Ga{\v{s}}i{\'c},
  Jur{\v{c}}{\'\i}{\v{c}}ek, Mairesse, Thomson, Yu, and
  Young}]{keizer2010parameter}
Simon Keizer, Milica Ga{\v{s}}i{\'c}, Filip Jur{\v{c}}{\'\i}{\v{c}}ek,
  Fran{\c{c}}ois Mairesse, Blaise Thomson, Kai Yu, and Steve Young. 2010.
\newblock Parameter estimation for agenda-based user simulation.
\newblock In \emph{11th Annual Meeting of the Special Interest Group on
  Discourse and Dialogue}, pages 116--123.

\bibitem[{Kottur et~al.(2017)Kottur, Moura, Lee, and Batra}]{kottur2017natural}
Satwik Kottur, Jos{\'e} Moura, Stefan Lee, and Dhruv Batra. 2017.
\newblock Natural language does not emerge ‘naturally’ in multi-agent
  dialog.
\newblock In \emph{2017 Conference on Empirical Methods in Natural Language
  Processing}, pages 2962--2967.

\bibitem[{Kreyssig et~al.(2018)Kreyssig, Casanueva, Budzianowski, and
  Gasic}]{kreyssig2018neural}
Florian Kreyssig, I{\~n}igo Casanueva, Pawe{\l} Budzianowski, and Milica Gasic.
  2018.
\newblock Neural user simulation for corpus-based policy optimisation of spoken
  dialogue systems.
\newblock In \emph{19th Annual Meeting of the Special Interest Group on
  Discourse and Dialogue}, pages 60--69.

\bibitem[{Lee et~al.(2019{\natexlab{a}})Lee, Cho, and
  Kiela}]{lee2019countering}
Jason Lee, Kyunghyun Cho, and Douwe Kiela. 2019{\natexlab{a}}.
\newblock Countering language drift via visual grounding.
\newblock In \emph{2019 Conference on Empirical Methods in Natural Language
  Processing and 9th International Joint Conference on Natural Language
  Processing}, pages 4376--4386.

\bibitem[{Lee et~al.(2019{\natexlab{b}})Lee, Zhu, Takanobu, Zhang, Zhang, Li,
  Li, Peng, Li, Huang, and Gao}]{lee2019convlab}
Sungjin Lee, Qi~Zhu, Ryuichi Takanobu, Zheng Zhang, Yaoqin Zhang, Xiang Li,
  Jinchao Li, Baolin Peng, Xiujun Li, Minlie Huang, and Jianfeng Gao.
  2019{\natexlab{b}}.
\newblock Convlab: Multi-domain end-to-end dialog system platform.
\newblock In \emph{57th Annual Meeting of the Association for Computational
  Linguistics: System Demonstrations}, pages 64--69.

\bibitem[{Lewis et~al.(2017)Lewis, Yarats, Dauphin, Parikh, and
  Batra}]{lewis2017deal}
Mike Lewis, Denis Yarats, Yann Dauphin, Devi Parikh, and Dhruv Batra. 2017.
\newblock Deal or no deal? end-to-end learning of negotiation dialogues.
\newblock In \emph{2017 Conference on Empirical Methods in Natural Language
  Processing}, pages 2443--2453.

\bibitem[{Li et~al.(2020)Li, Peng, Lee, Gao, Takanobu, Zhu, Huang, Schulz,
  Atkinson, and Adada}]{li2020results}
Jinchao Li, Baolin Peng, Sungjin Lee, Jianfeng Gao, Ryuichi Takanobu, Qi~Zhu,
  Minlie Huang, Hannes Schulz, Adam Atkinson, and Mahmoud Adada. 2020.
\newblock Results of the multi-domain task-completion dialog challenge.
\newblock In \emph{34th AAAI Conference on Artificial Intelligence, Eighth
  Dialog System Technology Challenge Workshop}.

\bibitem[{Liu and Lane(2017)}]{liu2017iterative}
Bing Liu and Ian Lane. 2017.
\newblock Iterative policy learning in end-to-end trainable task-oriented
  neural dialog models.
\newblock In \emph{2017 IEEE Automatic Speech Recognition and Understanding
  Workshop}, pages 482--489.

\bibitem[{Lowe et~al.(2017)Lowe, Wu, Tamar, Harb, Abbeel, and
  Mordatch}]{lowe2017multi}
Ryan Lowe, Yi~Wu, Aviv Tamar, Jean Harb, Pieter Abbeel, and Igor Mordatch.
  2017.
\newblock Multi-agent actor-critic for mixed cooperative-competitive
  environments.
\newblock In \emph{31st Annual Conference on Neural Information Processing
  Systems}, pages 6379--6390.

\bibitem[{Mnih et~al.(2015)Mnih, Kavukcuoglu, Silver, Rusu, Veness, Bellemare,
  Graves, Riedmiller, Fidjeland, Ostrovski, Petersen, Beattie, Sadik,
  Antonoglou, King, Kumaran, Wierstra, Legg, and Hassabis}]{mnih2015human}
Volodymyr Mnih, Koray Kavukcuoglu, David Silver, Andrei~A Rusu, Joel Veness,
  Marc~G Bellemare, Alex Graves, Martin Riedmiller, Andreas~K Fidjeland, Georg
  Ostrovski, Stig Petersen, Charles Beattie, Amir Sadik, Ioannis Antonoglou,
  Helen King, Dharshan Kumaran, Daan Wierstra, Shane Legg, and Demis Hassabis.
  2015.
\newblock Human-level control through deep reinforcement learning.
\newblock \emph{Nature}, 518(7540):529--533.

\bibitem[{Papangelis et~al.(2019)Papangelis, Wang, Molino, and
  Tur}]{papangelis2019collaborative}
Alexandros Papangelis, Yi-Chia Wang, Piero Molino, and Gokhan Tur. 2019.
\newblock Collaborative multi-agent dialogue model training via reinforcement
  learning.
\newblock In \emph{20th Annual Meeting of the Special Interest Group on
  Discourse and Dialogue}, pages 92--102.

\bibitem[{Peng et~al.(2017)Peng, Li, Li, Gao, Celikyilmaz, Lee, and
  Wong}]{peng2017composite}
Baolin Peng, Xiujun Li, Lihong Li, Jianfeng Gao, Asli Celikyilmaz, Sungjin Lee,
  and Kam-Fai Wong. 2017.
\newblock Composite task-completion dialogue policy learning via hierarchical
  deep reinforcement learning.
\newblock In \emph{2017 Conference on Empirical Methods in Natural Language
  Processing}, pages 2231--2240.

\bibitem[{Rashid et~al.(2018)Rashid, Samvelyan, Witt, Farquhar, Foerster, and
  Whiteson}]{rashid2018qmix}
Tabish Rashid, Mikayel Samvelyan, Christian~Schroeder Witt, Gregory Farquhar,
  Jakob Foerster, and Shimon Whiteson. 2018.
\newblock Qmix: Monotonic value function factorisation for deep multi-agent
  reinforcement learning.
\newblock In \emph{35th International Conference on Machine Learning}, pages
  4292--4301.

\bibitem[{Schatzmann et~al.(2007)Schatzmann, Thomson, Weilhammer, Ye, and
  Young}]{schatzmann2007agenda}
Jost Schatzmann, Blaise Thomson, Karl Weilhammer, Hui Ye, and Steve Young.
  2007.
\newblock Agenda-based user simulation for bootstrapping a pomdp dialogue
  system.
\newblock In \emph{2007 Conference of the North American Chapter of the
  Association for Computational Linguistics: Human Language Technologies},
  pages 149--152.

\bibitem[{Shah et~al.(2018)Shah, Hakkani-T{\"u}r, Liu, and
  T{\"u}r}]{shah2018bootstrapping}
Pararth Shah, Dilek Hakkani-T{\"u}r, Bing Liu, and Gokhan T{\"u}r. 2018.
\newblock Bootstrapping a neural conversational agent with dialogue self-play,
  crowdsourcing and on-line reinforcement learning.
\newblock In \emph{2018 Conference of the North American Chapter of the
  Association for Computational Linguistics: Human Language Technologies},
  pages 41--51.

\bibitem[{Shi et~al.(2019)Shi, Qian, Wang, and Yu}]{shi2019build}
Weiyan Shi, Kun Qian, Xuewei Wang, and Zhou Yu. 2019.
\newblock How to build user simulators to train rl-based dialog systems.
\newblock In \emph{2019 Conference on Empirical Methods in Natural Language
  Processing and 9th International Joint Conference on Natural Language
  Processing}, pages 1990--2000.

\bibitem[{Silver et~al.(2017)Silver, Schrittwieser, Simonyan, Antonoglou,
  Huang, Guez, Hubert, Baker, Lai, Bolton, Chen, Lillicrap, Hui, Sifre, Van
  Den~Driessche, Graepel, and Hassabis}]{silver2017mastering}
David Silver, Julian Schrittwieser, Karen Simonyan, Ioannis Antonoglou, Aja
  Huang, Arthur Guez, Thomas Hubert, Lucas Baker, Matthew Lai, Adrian Bolton,
  Yutian Chen, Timothy Lillicrap, Fan Hui, Laurent Sifre, George Van
  Den~Driessche, Thore Graepel, and Demis Hassabis. 2017.
\newblock Mastering the game of go without human knowledge.
\newblock \emph{Nature}, 550(7676):354--359.

\bibitem[{Stolcke et~al.(2000)Stolcke, Ries, Coccaro, Shriberg, Bates,
  Jurafsky, Taylor, Martin, Ess-Dykema, and Meteer}]{stolcke2000dialogue}
Andreas Stolcke, Klaus Ries, Noah Coccaro, Elizabeth Shriberg, Rebecca Bates,
  Daniel Jurafsky, Paul Taylor, Rachel Martin, Carol~Van Ess-Dykema, and Marie
  Meteer. 2000.
\newblock Dialogue act modeling for automatic tagging and recognition of
  conversational speech.
\newblock \emph{Computational linguistics}, 26(3):339--373.

\bibitem[{Su et~al.(2016)Su, Ga{\v{s}}i{\'c}, Mrk{\v{s}}i{\'c}, Barahona,
  Ultes, Vandyke, Wen, and Young}]{su2016line}
Pei-Hao Su, Milica Ga{\v{s}}i{\'c}, Nikola Mrk{\v{s}}i{\'c}, Lina M~Rojas
  Barahona, Stefan Ultes, David Vandyke, Tsung-Hsien Wen, and Steve Young.
  2016.
\newblock On-line active reward learning for policy optimisation in spoken
  dialogue systems.
\newblock In \emph{54th Annual Meeting of the Association for Computational
  Linguistics}, pages 2431--2441.

\bibitem[{Su et~al.(2018)Su, Li, Gao, Liu, and Chen}]{su2018discriminative}
Shang-Yu Su, Xiujun Li, Jianfeng Gao, Jingjing Liu, and Yun-Nung Chen. 2018.
\newblock Discriminative deep dyna-q: Robust planning for dialogue policy
  learning.
\newblock In \emph{2018 Conference on Empirical Methods in Natural Language
  Processing}, pages 3813--3823.

\bibitem[{Takanobu et~al.(2019)Takanobu, Zhu, and Huang}]{takanobu2019guided}
Ryuichi Takanobu, Hanlin Zhu, and Minlie Huang. 2019.
\newblock Guided dialog policy learning: Reward estimation for multi-domain
  task-oriented dialog.
\newblock In \emph{2019 Conference on Empirical Methods in Natural Language
  Processing and 9th International Joint Conference on Natural Language
  Processing}, pages 100--110.

\bibitem[{Van~Seijen et~al.(2017)Van~Seijen, Fatemi, Romoff, Laroche, Barnes,
  and Tsang}]{van2017hybrid}
Harm Van~Seijen, Mehdi Fatemi, Joshua Romoff, Romain Laroche, Tavian Barnes,
  and Jeffrey Tsang. 2017.
\newblock Hybrid reward architecture for reinforcement learning.
\newblock In \emph{31st Annual Conference on Neural Information Processing
  Systems}, pages 5392--5402.

\bibitem[{Williams et~al.(2016)Williams, Raux, and
  Henderson}]{williams2016dialog}
Jason~D Williams, Antoine Raux, and Matthew Henderson. 2016.
\newblock The dialog state tracking challenge series: A review.
\newblock \emph{Dialogue \& Discourse}, 7(3):4--33.

\bibitem[{Zhang et~al.(2019)Zhang, Li, Gao, and Chen}]{zhang2019budgeted}
Zhirui Zhang, Xiujun Li, Jianfeng Gao, and Enhong Chen. 2019.
\newblock Budgeted policy learning for task-oriented dialogue systems.
\newblock In \emph{57th Annual Meeting of the Association for Computational
  Linguistics}, pages 3742--–3751.

\bibitem[{Zhao et~al.(2019)Zhao, Xie, and Eskenazi}]{zhao2019rethinking}
Tiancheng Zhao, Kaige Xie, and Maxine Eskenazi. 2019.
\newblock Rethinking action spaces for reinforcement learning in end-to-end
  dialog agents with latent variable models.
\newblock In \emph{2019 Conference of the North American Chapter of the
  Association for Computational Linguistics: Human Language Technologies},
  pages 1208--1218.

\end{thebibliography}
\bibliographystyle{acl_natbib}

\newpage
\appendix

\section{Implementation Details}

Both the system policy $\pi$ and the user policy $\mu$ are implemented with two hidden layer MLPs. The action space of system policy and user policy is 172 and 80 respectively. For Hybrid Value Network $V$, all neural network units $f(\cdot)$ are two hidden layer MLPs. The activation function is all \textit{Relu} for MLPs.

We use RMSprop as the optimization algorithm. The batch size is set to 32. The weighted pretraining factor $\beta$ is 2.5, 4 for the system policy and user policy respectively. The learning rate for two polices is 1e-3 when pretraining. As for RL training, the learning rate is 1e-4, 5e-5 for the system policy and the user policy respectively, and 3e-5 for Hybrid Value Network. The discount factor $\gamma$ is 0.99, and the target network is updated every $C$= 400 training iterations.

In terms of reward design, the empty action penalty is set to -5, and penalties of other types are set to -1. The sub-goal completion reward is set to 5. The task success and the user goal reward are set to 20 if triggered, otherwise they are set to -5.

\section{Case Study}
To illustrate the superiority of our model, we show a pair of sample dialog sessions between the user and the system, trained with RL/RL and MADPL, respectively. As shown in Table \ref{tab:user_goal}, the user goal contains two domains (i.e., train and restaurant). The user agent is supposed to inform the constraints displayed in \textit{Info}, request information displayed in \textit{Reqt}, and express the intent to book the train and restaurant with the information in \textit{Book}. The system agent is required to answer the request, as well as book the train and the restaurant according to the user's demands.
Table \ref{tab:sample} shows the sampled dialog sessions. For RL/RL, the user sometimes talks about both the train and the restaurant simultaneously, making it difficult for the system to respond (highlighted in red color). Moreover, the system agent in RL/RL sometimes fails to answer the requested information or book the restaurant as the user demands (highlighted in blue color). As a result, the dialog in RL/RL fails as the user does not request the arrival time of the train, and the system does not book the restaurant for the user. In comparison, both the user and the system in MADPL fulfill their own task, and the whole dialog is successful.
Table \ref{tab:sample_da} shows the same sample dialog sessions with the original dialog acts.

\begin{table*}[!tb]
    \centering
    \small
\begin{tabular}{c|c|c|c|c|c}
\hline
\multicolumn{3}{c|}{\textbf{Train}} & \multicolumn{3}{c}{\textbf{Restaurant}} \\
\hline
Info & Reqt & Book & Info & Reqt & Book \\
\hline
destination=peterborough & duration & people= 3 & food=european & phone & time=19:30 \\
day=thursday & arriveBy &  & pricerange=expensive & postcode & people=5 \\
departure=cambridge & &  & & address & day=tuesday \\
leaveAt=19:45& &  & & & \\
\hline
\end{tabular}
    \caption{User goal used in the case study.}
    \label{tab:user_goal}
\end{table*}

\begin{table*}[!tb] 
\small
\begin{tabular}{l@{~}p{7cm}|l@{~}p{7cm}}
\hline
\multicolumn{2}{c|}{\textbf{RL/RL}}  & \multicolumn{2}{c}{\textbf{MADPL}} \\
\hline
U:& Hi, I'd like to take a train from cambridge to peterborough on thursday. I want to leave at about 19:45. & U:& Hi, I'd like to take a train from cambridge to peterborough on thursday. I want to leave at about 19:45.  \\
S:& You may consider TR7324, which leaves at 21:40. Shall I book tickets for you? & S:& You may consider TR7324, which leaves at 21:40, and arrives at cambridge at 24:23. Shall I book tickets for you?  \\
U:& Sounds good! Please book tickets to peterborough for 3 people. & U:& Sounds good! Please book tickets for 3 people.  \\
S:& I've booked your train tickets. The price is 10.10 pounds and your reference number is train-00002321. Is there anything else I can help you with today?  & S:& I 've booked your train tickets. The price is 10.10 pounds and your reference number is train-00002321. Is there anything else I can help you with today? \\
U:& \textcolor{red}{Could you please tell me the train ID and travel time? I also want to book an european restaurant with expensive price.} & U:& I also want to book an european restaurant with expensive price on Tuesday. \\
S:& There are 4 choices for expensive european restaurant. What area of town would you prefer ? & S:& There are 4 choices found. City Stop Restaurant is a good choice.  \\
U:& I don't care about the location. Could you please tell me the address, phone number and post of the restaurant? & U:& I don't care its name, but make sure it's an expensive restaurant. Can you help me with a reservation for 5 people at 19:30 this coming Tuesday? \\
S:& Sure, City Stop Restaurant is located at Cambridge City Football Club Milton Road Chesterton and the post is cb41uy. The phone number is 01223363270. Is there anything else I can help you with today? &  S:& Booking was successful. Reference number is: restaurant-00000083. \\
U:& \textcolor{blue}{Can you help me with a reservation for 5 people at 19:30 this coming Tuesday?} & U:& Although I don't care its name, I'd like to know its phone number and post.  \\
S:& \textcolor{blue}{Is there anything else I can help you with?} & S:& Sure, the phone number is 01223368786 and the post is cb23rh. Anything else? \\
U:& Could you please tell me the travel time of the train I just booked? &  U:& What's the address of the restaurant? \\
S:& The travel time is 60 minutes. Anything else I can look up for you? &  S:& The restaurant is located at Cambridge Lodge Hotel 139 Huntingdon Road City Centre. Is there anything else I can help you with?   \\
U:& \textcolor{blue}{Thanks! When will the train arrive?} & U:& Could you please tell me the travel time of the train I just booked? Thanks very much! \\
S:& \textcolor{blue}{You're welcome! Anything else? Bye!} & S:& The travel time is 105 minutes. Anything else I can look up for you? \\
U:& Thank you very much! & U:& Thank you. That 's all I needed . \\
S:& You are very welcome. Goodbye. & S:& You are very welcome. Goodbye. \\
\hline
\multicolumn{2}{c|}{\multirow[t]{2}{*}[1.3ex]{\textbf{Failure:}} \begin{tabular}[c]{@{}p{6.5cm}@{}}User does not request the arrive time of the train, and the system does not book the restaurant for the user.\end{tabular}} & \multicolumn{2}{c}{\textbf{Success}}
\\ \hline
\end{tabular}
\small
\centering
\caption{A sample dialog session comparison between RL/RL and MADPL in natural language.}
\label{tab:sample}
\end{table*}

\begin{table*}[!tb] 
\small
\begin{tabular}{l@{~}p{6.5cm}|l@{~}p{6.5cm}}
\hline
\multicolumn{2}{c|}{\textbf{RL/RL}}  & \multicolumn{2}{c}{\textbf{MADPL}} \\
\hline
U:& \{'train-inform-dest': 'peterborough', 'train-inform-day': 'thursday', 'train-inform-depart': 'cambridge', 'train-inform-leave': '19:45'\} & U:& \{'train-inform-dest': 'peterborough', 'train-inform-day': 'thursday', 'train-inform-depart': 'cambridge', 'train-inform-leave': '19:45'\}  \\
S:& \{'train-inform-leave': '21:40', 'train-inform-id': 'TR7324', 'train-offerbook-none': 'none'\} & S:& \{'train-inform-leave': '21:40', 'train-inform-id': 'TR7324', 'train-inform-arrive': '24:23', 'train-offerbook-none': 'none', 'train-inform-dest': 'cambridge'\}  \\
U:&\{'train-inform-dest': 'peterborough', 'train-inform-people': '3'\} & U:& \{'train-inform-people': '3'\}  \\
S:& \{'general-reqmore-none': 'none', 'train-offerbooked-ref': 'train-00002321', 'train-offerbooked-ticket': '10.10 pounds'\}  & S:& \{'general-reqmore-none': 'none', 'train-offerbooked-ref': 'train-00002321', 'train-offerbooked-ticket': '10.10 pounds'\} \\
U:& \textcolor{red}{\{'restaurant-inform-food': 'european', 'restaurant-inform-price': 'expensive', 'train-request-time': '?', 'train-request-id': '?'\}} & U:& \{'restaurant-inform-food': 'european', 'restaurant-inform-price': 'expensive', 'restaurant-inform-day': 'tuesday'\} \\
S:& \{'restaurant-inform-choice': '4', 'restaurant-inform-food': 'european', 'restaurant-inform-price': 'expensive', 'restaurant-request-area-?': '?'\} & S:& \{'restaurant-inform-choice': '4', 'restaurant-recommend-name': 'city stop restaurant'\}  \\
U:& \{'restaurant-inform-area': 'dont care', 'restaurant-request-addr': '?', 'restaurant-request-phone': '?', 'restaurant-request-post': '?'\} & U:& \{'restaurant-inform-price': 'expensive', 'restaurant-inform-time': '19:30', 'restaurant-inform-day': 'tuesday', 'restaurant-inform-people': '5', 'restaurant-inform-name': 'dont care'\} \\
S:& \{'general-reqmore-none': 'none', 'restaurant-inform-name': 'city stop restaurant', 'restaurant-inform-addr': 'Cambridge City Football Club Milton Road Chesterton', 'restaurant-recommend-name': 'city stop restaurant', 'restaurant-inform-phone': '01223363270', 'restaurant-inform-post': 'cb41uy'\} &  S:& \{'general-reqmore-none': 'none', 'booking-book-ref': 'restaurant-00000083'\} \\
U:& \textcolor{blue}{\{'restaurant-inform-time': '19:30', 'restaurant-inform-day': 'tuesday', 'restaurant-inform-people': '5'\}} & U:& \{'restaurant-inform-name': 'dont care', 'restaurant-request-phone': '?', 'restaurant-request-post': '?'\}  \\
S:& \textcolor{blue}{\{'general-reqmore-none': 'none'\}} & S:& \{'general-reqmore-none': 'none', 'restaurant-inform-phone': '01223368786', 'restaurant-inform-post': 'cb23rh'\} \\
U:& \{'train-request-time': '?'\} &  U:& \{'restaurant-request-addr': '?'\} \\
S:& \{'general-reqmore-none': 'none', 'train-inform-time': '60 minutes'\} &  S:& \{'general-reqmore-none': 'none', 'restaurant-inform-addr': 'Cambridge Lodge Hotel 139 Huntingdon Road City Centre'\}  \\
U:& \textcolor{blue}{\{'general-thank-none': 'none', 'train-request-arrive': '?'\}} & U:& \{'general-thank-none': 'none', 'train-request-time': '?'\} \\
S:& \textcolor{blue}{\{'general-reqmore-none': 'none', 'general-bye-none': 'none', 'general-welcome-none': 'none'\}} & S:& \{'general-reqmore-none': 'none', 'train-inform-time': '105 minutes'\} \\
U:& \{'general-thank-none': 'none'\} & U:& \{'general-thank-none': 'none'\} \\
S:& \{'general-bye-none': 'none', 'general-welcome-none': 'none'\} & S:& \{'general-bye-none': 'none', 'general-welcome-none': 'none'\} \\
\hline
\multicolumn{2}{c|}{\multirow[t]{2}{*}[1.3ex]{\textbf{Failure:}} \begin{tabular}[c]{@{}p{6.5cm}@{}}User does not request the arrive time of the train, and the system does not book the restaurant for the user.\end{tabular}} & \multicolumn{2}{c}{\textbf{Success}}
\\ \hline
\end{tabular}
\small
\centering
\caption{A sample dialog session comparison between RL/RL and MADPL in dialog acts.}
\label{tab:sample_da}
\end{table*}

\end{document}